\documentclass[letterpaper]{article} 
\usepackage{aaai24}  
\usepackage{times}  
\usepackage{helvet}  
\usepackage{courier}  
\usepackage[hyphens]{url}  
\usepackage{graphicx} 
\usepackage{natbib}  
\usepackage{caption} 
\usepackage{algorithm}
\usepackage{algorithmicx}  
\usepackage{algpseudocodex} 

\usepackage{amsmath}
\usepackage{amssymb}
\usepackage{mathtools}
\usepackage{bm}
\usepackage{multirow}
\usepackage{diagbox}
\usepackage{dsfont}

\usepackage{booktabs}

\usepackage{newfloat}
\usepackage{listings}

\newtheorem{prop}{Proposition}
\newtheorem{definition}{Definition}
\newtheorem{ass}{Assumption}
\newtheorem{cor}{Corollary}
\newtheorem{remark}{Remark}

\newenvironment{proof}{{\noindent\it Proof}\quad}{\hfill $\square$\par}
\DeclareCaptionStyle{ruled}{labelfont=normalfont,labelsep=colon,strut=off} 
\floatstyle{ruled}
\newfloat{listing}{tb}{lst}{}
\floatname{listing}{Listing}
\title{Regroup Median Loss for Combating Label Noise}
\author{
    Fengpeng Li\textsuperscript{\rm 1}, 
    Kemou Li\textsuperscript{\rm 1}, 
    Jinyu Tian\textsuperscript{\rm 2},
    Jiantao Zhou\textsuperscript{\rm 1}\thanks{Corresponding author.}
}
\affiliations{
    \textsuperscript{\rm 1}State Key Laboratory of Internet of Things for Smart City,\\
    Department of Computer and Information Science, University of Macau\\
    \textsuperscript{\rm 2}Faculty of Innovation Engineering, Macau University of Science and Technology\\
    \{yc17420, mc15415, jtzhou\}@um.edu.mo, jytian@must.edu.mo

}

\usepackage{bibentry}

\begin{document}

\maketitle

\begin{abstract}
The deep model training procedure requires large-scale datasets of annotated data. Due to the difficulty of annotating a large number of samples, label noise caused by incorrect annotations is inevitable, resulting in low model performance and poor model generalization. To combat label noise, current methods usually select clean samples based on the small-loss criterion and use these samples for training. Due to some noisy samples similar to clean ones, these small-loss criterion-based methods are still affected by label noise. To address this issue, in this work, we propose Regroup Median Loss (RML) to reduce the probability of selecting noisy samples and correct losses of noisy samples. RML randomly selects samples with the same label as the training samples based on a new loss processing method. Then, we combine the stable mean loss and the robust median loss through a proposed regrouping strategy to obtain robust loss estimation for noisy samples. To further improve the model performance against label noise, we propose a new sample selection strategy and build a semi-supervised method based on RML. Compared to state-of-the-art methods, for both the traditionally trained and semi-supervised models, RML achieves a significant improvement on synthetic and complex real-world datasets. The source code of the paper has been released.
\end{abstract}

\section{Introduction}
\label{intro}
Deep learning model has been proven powerful in various practical tasks, \emph{e.g.}, image classification \cite{szegedy2015going,zhang2015accelerating,krizhevsky2017imagenet}, object detection \cite{girshick2015fast,redmon2016you} and image semantic segmentation \cite{he2017mask,zhang2018context}. As is known, the performance of deep learning models heavily relies on dataset scale and annotation quality. Currently, collecting large-scale datasets with high-quality annotation is an extremely expensive task. An efficient and cheap method called crowd-sourcing policy is often used to collect large-scale datasets \cite{zubiaga2015crowdsourcing}. However, in the crowd-sourcing data labeling procedure, it is inevitable that some samples will be annotated with incorrect labels, resulting in the so-called label noise. Due to the strong capacity of the deep learning model, it can easily fit the training samples with incorrect labels, which impairs the performance and generalization of the deep model \cite{zhang2017understanding}. Owing to the difficulty of selecting noisy samples in a large-scale dataset, how to make the deep model robust to label noise becomes an essential and fundamental research topic \cite{bai2021understanding,karim2022unicon}.

To protect the deep model from label noise \cite{DBLP:conf/nips/HanYYNXHTS18,bai2021understanding,karim2022unicon}, some methods have been proposed recently. According to the operations of these methods, they can be divided into two categories: loss correction and sample selection \cite{karim2022unicon}. Loss correction commonly attempts to estimate the noise transition matrix. The matrix contains the conditional probabilities of samples belonging to different classes based on their observed labels \cite{frenay2013classification}. However, estimating the transition matrix is difficult especially when the number of classes is large and the ratio of noisy samples is too high \cite{xu2019dmi,karim2022unicon}. Compared to loss correction, sample selection methods pay more attention to filtering out samples with incorrect labels based on the small-loss criterion \cite{DBLP:conf/nips/HanYYNXHTS18,yu2019does}, supposing clean samples have smaller losses than noisy ones. Inevitably, some noisy samples will also have small loss values, and will therefore be misclassified as clean. Traditional small-loss criterion based methods usually select only clean samples for training and discard noisy ones \cite{DBLP:conf/nips/HanYYNXHTS18, xia2019anchor}, resulting in information loss. The semi-supervised methods \cite{li2020dividemix,karim2022unicon} use the small-loss criterion to select clean samples and relabel noisy ones. However, these methods require complex strategies to separate clean and noisy samples, which increases time consumption.

\begin{figure*}
    \centering
    \includegraphics[scale=0.35]{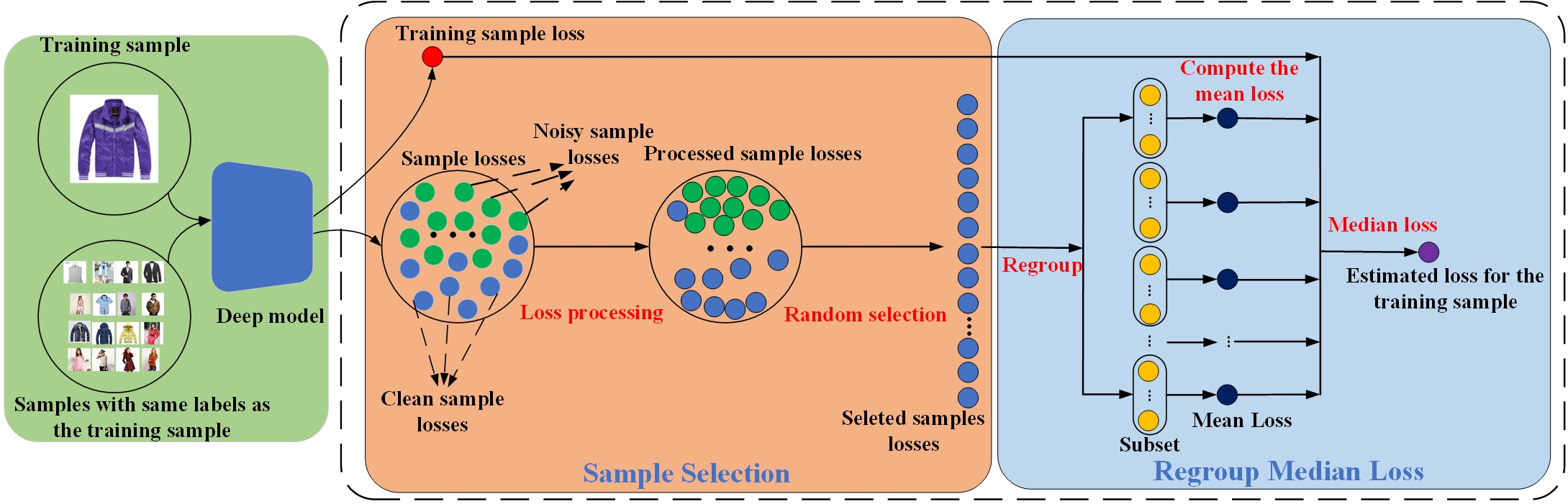}
    \caption{The overview of RML, which consists of two steps: (\romannumeral1) Sample Selection and (\romannumeral2) Regroup Median Loss.}
    \label{step}
\end{figure*}

According to \cite{xia2020part}, noisy samples can also provide useful information for model training if their losses can be appropriately corrected. Motivated by these previous works, we propose Regroup Median Loss (RML) to reduce the probability of selecting noisy samples and correct the distorted losses based on losses of clean samples through a robust estimation method. In RML, we select samples with the same observed label as each training sample and use their losses to estimate the loss of the training sample. To reduce the probability of selecting noisy samples, we propose a new strategy to process losses and use them to select samples. However, noisy samples may still be selected inevitably. Loss estimation by the mean loss of all selected samples will be influenced. According to \cite{nemirovskij1983problem, catoni2012challenging}, Median-of-Means (MoM) approach is particularly suitable for heavy-tailed distributions and is insensitive to outliers. Motivated by MoM, we propose a robust loss estimation method that combines median loss and mean loss. During the loss estimation procedure, our proposed RML divides the selected samples into several groups and computes the mean loss of each group. Then, we compute the median between these mean losses and the training sample loss to robustly estimate the training sample loss. In addition, we build a semi-supervised method based on RML to combat label noise. 
 
 The contributions of this paper are presented as follows:
 \begin{itemize}

    \item We propose a loss processing method to reduce the probability of selecting noisy samples. We theoretically explain and verify the operation that reduces the probability of selecting noisy samples.
    
     \item We propose Regroup Median Loss, a novel robust loss estimation method that can correct the distorted loss value of noisy samples. Then these corrected losses of training samples can mitigate the negative impact of noisy samples and improve the model performance.

     \item We verify that RML can achieve better model performance than existing methods on synthetic and real-world datasets with various types of label noises. The proposed method achieves about 6\% performance improvement on CIFAR-100 with various types of label noises and increases the  test accuracy by about 8\% on the challenging real-world dataset WebVision. Moreover, the semi-supervised method based on RML achieves state-of-the-art performance on synthetic and real-world datasets.
 \end{itemize}

\section{Related Works}
 This section summarizes existing methods to combat label noise. To better understand the motivation of existing methods, we first introduce the types of label noises. According to \cite{DBLP:journals/corr/abs-2007-08199}, label noise can be divided into two categories: instance-independent label noise and instance-dependent label noise. For instance-independent label noise, it is only relevant to the original sample labels and independent of the sample features \cite{van2015learning}. Symmetric label noise and asymmetric label noise are two typical types of instance-independent label noise. Compared to instance-independent label noise, the generating corruption probability of instance-dependent label noise depends on both sample features and labels \cite{xia2020part}. Some studies show that instance-dependent label noise is common in real-world datasets, such as Clothing1M \cite{xiao2015learning} and WebVision \cite{li2017WebVision}. 

Due to the negative impact of label noise on model generalization, a large number of methods have been proposed \cite{DBLP:conf/nips/HanYYNXHTS18,yu2019does,DBLP:conf/nips/BerthelotCGPOR19,bai2021understanding,karim2022unicon}. They can be roughly divided into two main species: loss correction and sample selection. Loss correction adopts the noise transition matrix to weigh label smoothing parameters \cite{xia2021sample,DBLP:conf/iclr/ChengZLGSL21}. The difficulty of this method is how to obtain the noise transition matrix. One commonly used method tends to count the transition relationship between clean dataset labels and noisy dataset labels \cite{xia2019anchor,DBLP:conf/iclr/ChengZLGSL21,zhu2021second} and use the label change frequency as the transition probability to build the true noise transition matrix. The other method attempts to use some constraints or characteristics of the noise transition matrix, such as total variation \cite{zhang2021learning} and diagonal dominance \cite{li2021provably} to obtain a transition matrix by gradient descent. Although the matrix obtained by the former method has a remarkable performance against label noise than the calculated transition matrix, it is difficult to obtain the original clean dataset in a practical task \cite{karim2022unicon}. In addition, label smoothing \cite{muller2019does} is also a way to achieve loss correction for noisy data \cite{zhang2021delving,cui2022improving}.

Sample selection is another type of method that aims to train models on small-loss samples. Intuitively, small-loss samples are more likely to be correctly labeled. The memorization effect of DNNs shows that even with noisy labels, DNNs learn clean and simple patterns first, and then gradually fit all samples \cite{10.5555/3305381.3305406}. This has given rise to the widely used small-loss criterion: considering small-loss samples as clean ones. Co-teaching \cite{DBLP:conf/nips/HanYYNXHTS18} is a typical method based on the small-loss criterion with two deep models. However, the selection operation based on the small-loss criterion can still make mistakes, misclassifying noisy samples as clean ones, and vice versa. To further address the shortcomings of the traditional small-loss criterion based methods, a semi-supervised method, DivideMix \cite{li2020dividemix}, adopts Gaussian Model Mixture  \cite{rasmussen1999infinite} method to select clean samples based on sample losses and uses MixMatch \cite{DBLP:conf/nips/BerthelotCGPOR19} to train two deep models. Although DivideMix has remarkable performance against label noise, it requires the noise rate of the dataset, which significantly limits its application. To further address the issue of DivideMix, some methods, such as MOIT \cite{ortego2021multi}, Jo-SRC \cite{yao2021jo} and UNICON \cite{karim2022unicon}, develop new strategies to separate clean and noisy samples.

\section{Methods}
In this section, we present the details of our proposed approach for combating label noise. We start by introducing some fundamental notations, followed by a two-part description of the RML. We then analyze the robustness of RML theoretically. Lastly, we introduce a common method and a semi-supervised model based on RML.

\subsection{Notations} 
We use bold capital letters such as $\bm{X}$ to represent a random vector, bold lowercase letters such as $\bm{x}$ to represent the realization of a random vector, capital letters such as $Y$ to represent a random variable, and lowercase letters such as $y$ to represent the realization of a random variable. Consider a $c$-class classification problem, let $\mathcal{X}$ be the feature space, $\mathcal{Y}=\{1,\dots,c\}$ be the label space, $(\bm{X},Y)\in\mathcal{X}\times\mathcal{Y}$ be the random variables with joint distribution $P_{\bm{X},Y}$ and $\mathcal{D}=\left\{(\bm{x}_i,{y}_i)\right\}^N_{i=1}$ be a dataset containing i.i.d. $N$ samples drawn from $P_{\bm{X},Y}$,  where $\bm{x}_i$ and $y_i$ are the $i$-th instance and its label. In practical applications, the true label $Y$ may not be observable. Instead, we have an noisy dataset $\tilde{\mathcal{D}}=\left\{(\bm{x}_i,\tilde{y}_i)\right\}^N_{i=1}$ consisting of i.i.d. $N$ samples drawn from $P_{\bm{X},\tilde{Y}}$, where $\bm{x}_i$ is the $i$-th instance and $\tilde{y}_i$ is its observed label which may be correct or not. Define the loss set of $\tilde{\mathcal{D}}$ as $\left.\mathcal{L}(\tilde{\mathcal{D}})\coloneqq\big\{\ell(f(\bm{x}_i),\tilde{y}_i)\mid(\bm{x}_i,\tilde{y}_i)\in\tilde{\mathcal{D}}\big\}^N_{i=1}\right.$, where $\ell(\cdot)$ denotes the cross-entropy (CE) loss function and $f$ is a deep model with parameter $\theta\in \mathbb{R}^d$.

\begin{figure}
    \centering
    \includegraphics[scale=0.265]{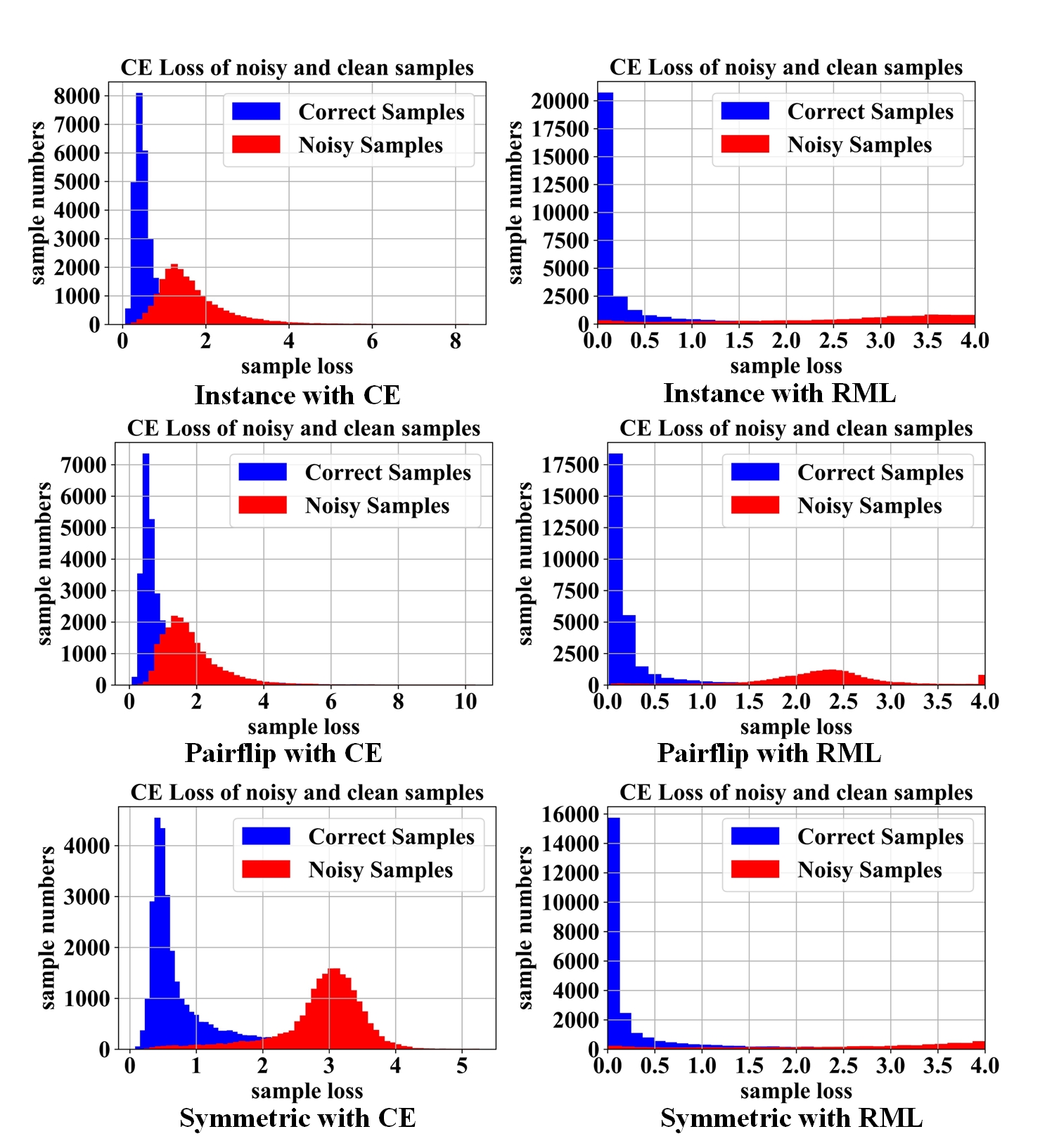}
    \caption{Loss distribution of noisy and clean samples on CIFAR-10 for the model trained with CE and RML. }
    \label{fig2}
\end{figure}

\subsection{Regroup Median Loss}
Fig. \ref{step} illustrates the two-step process of RML. In Step (\romannumeral1), a newly-proposed loss processing method is used to select clean samples. Step (\romannumeral2) utilizes losses of the selected samples and the original sample to estimate the loss by a regroup strategy. Next we provide more details on these two steps.

\paragraph{Sample Selection.} In the process of RML, first of all, for each training sample, we select samples with the same observed labels as this given sample. Consider a training sample $(\bm{x},\tilde{y})$ in $\tilde{\mathcal{D}}$, suppose that there are totally $m$ samples with the same label as $\tilde{y}$ in $\tilde{\mathcal{D}}$. We denote by $\left.\tilde{\mathcal{D}}_{\tilde{y}}=\big\{(\bm{x}_{i},\tilde{y}_{i})\in\tilde{\mathcal{D}}\mid \tilde{y}_{i}=\tilde{y}\big\}^{m}_{i=1}=\left\{(\bm{x}_{i},\tilde{y})\right\}^{m}_{i=1}\right.$ the selected set and $\left.\mathcal{L}(\tilde{\mathcal{D}}_{\tilde{y}})=\big\{\ell(f(\bm{x}_i), \tilde{y})\mid(\bm{x}_i,\tilde{y})\in\tilde{\mathcal{D}}_{\tilde{y}}\big\}^m_{i=1}\right.$ $=\left\{\ell_i \right\}^m_{i=1}$ the loss set of $\tilde{\mathcal{D}}_{\tilde{y}}$. Since clean samples usually have small losses \cite{DBLP:conf/ijcai/GuiWT21}, in order to select samples with small losses from $\tilde{\mathcal{D}}_{\tilde{y}}$, we define a new sample selection strategy based on sample loss.
\begin{definition}\label{def1}
  For any $(\bm{x}_{i},\tilde{y}_{i})\in\tilde{\mathcal{D}}_{\tilde{y}}$, define its selection probability $p_i$ based on its original CE loss $\ell_i$ as 
\begin{equation}
\label{prob1}
    p_i\coloneqq\frac{\mathrm{e}^{-\ell_i}}{\sum_{j=1}^{m}\mathrm{e}^{-\ell_j}}.
\end{equation}
\end{definition}

By Def. \ref{def1}, the larger the sample loss, the smaller its selection probability, \emph{i.e.}, clean samples have a higher selection probability. Nevertheless, such sample selection based on original CE loss still has the same issue as the current small-loss criterion because of the loss overlap between the clean and noisy samples \cite{bai2021understanding}. To better separate noise and clean samples, we processed the sample loss. 
\begin{definition}
    For any $(\bm{x}_{i},\tilde{y}_{i})\in\tilde{\mathcal{D}}_{\tilde{y}}$, define its processed loss $\tilde{\ell}_i$ based on its original CE loss $\ell_i$ as
\begin{equation}
\label{processed}
    \tilde{\ell}_i\coloneqq\ell_i\times(\ell_i+\varepsilon),
\end{equation}
where $\varepsilon$ is a bias term. Thus, its improved selection probability $\tilde{p}_i$ based on its processed loss $\tilde{\ell}_i$ is
\begin{equation}
\label{prob}
    \tilde{p}_i\coloneqq\frac{\mathrm{e}^{-\tilde\ell_i}}{\sum_{j=1}^{m}\mathrm{e}^{-\tilde\ell_j}}.
\end{equation}
\end{definition}

Then, we employ Prop. \ref{prop1} to explain the operation.


\begin{prop}\label{prop1}{For an arbitrary sample $(\bm{x}_{\tau},\tilde{y})$ in $\tilde{\mathcal{D}}_{\tilde{y}}$, after the processing operation in Eq. \eqref{processed}, its selection probability change is $\log p_{\tau}-\log \tilde{p}_{\tau}=\ell_{\tau}(\ell_{\tau}+\varepsilon-1)-\beta=\ell_{\tau}^2-\beta$, where $\beta=\log\frac{\sum^m_{j=1}\mathrm{e}^{-\ell_j}}{\sum^m_{j=1}\mathrm{e}^{-\ell_j (\ell_j +\varepsilon)}}$ is a constant.}
\end{prop}


Prop. \ref{prop1} demonstrates the change of selection probability between Eq. \eqref{prob1} and \eqref{prob}. Since $\sum^m_{j=1}\mathrm{e}^{-\ell_j (\ell_j +\varepsilon)}=\left.\sum^m_{j=1}\mathrm{e}^{-\ell_j ^2 -\ell_j}<\sum^m_{j=1}\mathrm{e}^{-\ell_j}\right.$, then $\beta>0$ and has a fixed value for each training epoch. If $\ell_{\tau}^2>\beta$, which means the loss value of the sample is large enough, then $\log p_{\tau}>\log \tilde{p}_{\tau}$ and thus $\left.\tilde{p}_{\tau}<p_{\tau}\right.$, the selection probability is reduced. On the contrary, the selection probability is improved when $\ell_{\tau}^2<\beta$, indicating the sample has small loss. Therefore, the processing operation in Eq. \eqref{processed} reduces the selection probability of noisy samples while increasing that of clean ones. Notably, the setting of $\varepsilon$ affects the number of samples with high selection probability. The larger $\varepsilon$ is, the fewer samples with high selection probability, resulting in too few samples to select and overfitting issues. Setting $\varepsilon=1$ can achieve a trade-off between high selection probability of clean samples and diversity of selected samples.

\paragraph{Regroup Median Loss.} 
As stated above, the proposed method reduces the probability of selecting noisy samples. The samples selected by the probability according to Eq. \eqref{prob} are more likely to have correct labels. To make full use of the information about noisy samples, we then use the selected samples to estimate the loss of training samples. 

First of all, for a training sample $(\bm{x}, \tilde{y})$ in a mini-batch, recall that RML records losses of all samples in $\tilde{\mathcal{D}}_{\tilde{y}}$ as $\mathcal{L}(\tilde{\mathcal{D}}_{\tilde{y}})$. Then we randomly select $n\times k$ samples according to Eq. \eqref{prob} without replacement, and combine them as $\tilde{\mathcal{D}}^{s}_{\tilde{y}}=\left\{(\bm{x^s}_i, \tilde{y})\right\}_{i=1}^{n\times k}\subseteq\tilde{\mathcal{D}}_{\tilde{y}}$, where $\bm{x}_i^s$ is the $i$-th selected sample, $n\times k\leqslant m$ and $n$ is an even number. $\tilde{\mathcal{D}}^{s}_{\tilde{y}}$ follows a distribution $P_{\bm{X^s}|\tilde{Y}=\tilde{y}}$ whose p.d.f. is given by Eq. \eqref{prob}. After the selection operation, these selected samples are used to estimate the training sample loss. 

Although the median loss of samples in $\tilde{\mathcal{D}}_{\tilde{y}}$ can mitigate the negative impact of outliers of the loss distribution of selected samples, it is small and has a large fluctuation between different training steps \cite{staerman2021ot}, resulting in the difficulty of the model training procedure convergence. Compared to the median loss, the mean loss is stable, although it is easily affected by the distorted loss of noisy samples. To obtain a stable and robust loss estimation, RML combines the mean loss and the median loss. RML partitions $\tilde{\mathcal{D}}^{s}_{\tilde{y}}$ into $n$ disjoint subsets $\left.\mathcal{S}_1,\dots,\mathcal{S}_n\right.$ of size $\left.|\mathcal{S}_i|=k\right.$ as $\left.\tilde{\mathcal{D}}^{s}_{\tilde{y}}=\bigcup_{i=1}^{n}\mathcal{S}_{i}\right.$. Then we calculate the mean loss of each subset $\mathcal{S}_i$ and store them as $\left.\mathcal{W}=\big\{\ell^s_i\mid\ell^s_i=\frac{1}{|\mathcal{S}_i|}\sum_{(\bm{x^s},\tilde{y})\in\mathcal{S}_i}\ell(f(\bm{x^s}),\tilde{y})\big\}^n_{i=1}\right.$. For a single training sample $(\bm{x},\tilde{y})$, RML estimates its loss by the median of $\mathcal{W}$ and its original loss $\ell(f(\bm{x}),\tilde{y})$ as
\begin{equation}
\label{esti}
    \ell_{\mathrm{RML}}(f(\bm{x}),\tilde{y})\coloneqq\mathrm{Median}\big(\mathcal{W}\cup\{\ell(f(\bm{x}),\tilde{y})\}\big).
\end{equation}

Intuitively, if the observed label $\tilde{y}$ is the true label of $\bm{x}$, such an operation ensures that the estimated loss does not deviate too far from its true loss. Specifically, selecting samples randomly makes each subset of $\tilde{\mathcal{D}}^{s}_{\tilde{y}}$ contains different samples in different training steps, thus protecting the model from overfitting issues caused by the fixed sample combination. Compared to the Median-of-Means, using the mean of the median losses of each subset can also mitigate the negative of selected noisy samples in $\tilde{\mathcal{D}}^{s}_{\tilde{y}}$. However, when the observed label $\tilde{y}$ is incorrect, $\ell_{\mathrm{RML}}(f(\bm{x}),\tilde{y})$ obtained by the mean of median losses of all subsets is distorted. To make the model training procedure with RML efficient, we add some operations, which can be found in Appendix \ref{details}.

\begin{table*}[htbp]

\small
\centering

\scalebox{1}{
\begin{tabular}{c|c|c|c|c|c|c}
\hline

\multirow{2}{*}{Dataset} & \multirow{2}*{Method} & \multicolumn{2}{c|}{Symmetric} & Pairflip & \multicolumn{2}{c}{Instance} \\
\cline{3-7}& &0.2&0.5&0.45&0.2&0.4\\ \hline
\multirow{8}{*}{CIFAR-10}      & CE  &   84.00$\pm$0.66        &  75.51$\pm$1.24          &    63.34$\pm$6.03          &      85.10$\pm$0.68      &    77.00$\pm$2.17                     \\  
& Co-teaching  &       87.16$\pm$0.11      &  72.80$\pm$0.45          &  70.11$\pm$1.16           & 86.54$\pm$0.11          & 80.98$\pm$0.39                      \\                         
      & Forward           & 85.63$\pm$0.52           & 77.92$\pm$0.66            &      60.15$\pm$1.97       & 85.29$\pm$0.38 & 74.72$\pm$3.24                       \\                    
 & Joint-Optim           & 89.70$\pm$0.11            &  85.00$\pm$0.17            & 82.63$\pm$1.38            & 89.69$\pm$0.42 & 82.62$\pm$0.57                       \\  
 & T-revision           & 89.63$\pm$0.13            & 83.40$\pm$0.65           &  77.06$\pm$6.47           &  90.46$\pm$0.13   & 85.37$\pm$3.36                    \\ 
 & DMI           &   88.18$\pm$0.36          &  78.28$\pm$0.48  & 57.60$\pm$14.56            & 89.14$\pm$0.36     & 84.78$\pm$1.97                  \\ 
  & PES           &  92.38$\pm$0.40          & 87.45$\pm$0.35           & 88.43$\pm$1.08           & 92.69$\pm$0.44  & 89.73$\pm$0.51                    \\ 
 
  & RML (Ours) & \textbf{92.89$\pm$0.42}          &  \textbf{89.13$\pm$0.21}           &  \textbf{90.54$\pm$0.34}          &   \textbf{93.34$\pm$0.23}  &\textbf{90.73$\pm$0.44}                    \\ 
                        
                        \hline 
                         \hline

\multirow{8}{*}{CIFAR-100}      & CE  &  51.43$\pm$0.58           &  37.69$\pm$3.45          &   34.10$\pm$2.04          &  52.19$\pm$1.42          &       42.26$\pm$1.29                \\  
& Co-teaching  &   59.28$\pm$0.47         & 41.37$\pm$0.08           &   33.22$\pm$0.48         &  57.24$\pm$0.69 & 45.69$\pm$0.99                              \\  
& Forward  &    57.75$\pm$0.37          &   44.66$\pm$1.01          &   27.88$\pm$0.80          & 58.76$\pm$0.66           &  44.50$\pm$0.72                      \\ 
& Joint-Optim  &    64.55$\pm$0.38          & 50.22$\pm$0.41            &  42.61$\pm$0.61            &  65.15$\pm$0.31           &      55.57$\pm$0.41                  \\ 
& T-revision  &     65.40$\pm$1.07         &  50.24$\pm$1.45           &  41.10$\pm$1.95            &  60.71$\pm$0.73           &    51.54$\pm$0.91                    \\ 
& DMI  &   58.73$\pm$0.70           &  44.25$\pm$1.14           &   26.90$\pm$0.45           & 58.05$\pm$0.20            &          47.36$\pm$0.68             \\ 
& PES  &  68.89$\pm$0.45            &  58.90$\pm$2.72           &  57.18$\pm$1.44            & 70.49$\pm$0.79            &            65.68$\pm$1.41            \\ 
& RML (Ours)   & \textbf{ 69.74$\pm$0.24 }           & \textbf{ 67.05$\pm$0.31 }          &   \textbf{ 65.67$\pm$0.54 }         & \textbf{ 73.43$\pm$0.71 }           & \textbf{ 71.12$\pm$1.12 }                      \\ \hline

\end{tabular}
}

\normalsize
\caption{Average test accuracy ($\%$) comparison with state-of-the-art methods without semi-supervised strategy on CIFAR-10
and CIFAR-100. The mean and standard deviation computed over five runs are presented. Baselines are from \cite{bai2021understanding}.}
\label{unsemisym}
\end{table*}

\subsection{Robustness Analysis of RML }

This section analyzes the robustness of Eq. \eqref{esti}, which is our main objective. Note that $\ell(f(\bm{x}),\tilde{y})$ can also be viewed as the mean loss of a set. Thus, we consider a set $\mathcal{S}_{n+1}$ containing $k$ samples drawn from $P_{\bm{X^s}|\tilde{Y}=\tilde{y}}$ and satisfying $\left.\frac{1}{|\mathcal{S}_{n+1}|}\sum_{(\bm{x^s},\tilde{y})\in\mathcal{S}_{n+1}}\ell(f(\bm{x^s}),\tilde{y})=\ell(f(\bm{x}),\tilde{y})\right.$. According to \cite{10.1214/19-AOS1828}, such RML can be regarded as an MoM estimator of the loss expectation of $\tilde{\mathcal{D}}_{\tilde{y}}^{*}=\bigcup_{i=1}^{n+1}\mathcal{S}_{i}$. To introduce the property of RML, we need to make some mild assumptions about $\tilde{\mathcal{D}}_{\tilde{y}}^{*}$. 
\begin{ass}
    Assume that the samples in $\tilde{\mathcal{D}}_{\tilde{y}}^{*}$ are i.i.d. drawn from a distribution $P_{\bm{X^*}|\tilde{Y}=\tilde{y}}$, satisfying $\mathbb{E}_{\bm{X^*}|\tilde{Y}=\tilde{y}}[\ell(f(\bm{X^{*}}),\tilde{y})]=\hat{\mu}$ and $\mathrm{Var}_{\bm{X^*}|\tilde{Y}=\tilde{y}}[\ell(f(\bm{X^{*}}),\tilde{y})]$ $=\hat{\sigma}^2<\infty$.
\end{ass}

Then based on the properties of MoM, we employ Prop. \ref{prop2} to prove the robustness of $\ell_{\mathrm{RML}}$. The proof of the proposition is in Appendix \ref{proof}.
\begin{prop}\label{prop2}{For any $\left. n, \epsilon > 0 \right.$, $\ell_{\mathrm{RML}}(f(\bm{x}),\tilde{y})$ satisfies $\left.\mathbb{P}\left(|\ell_{\mathrm{RML}}(f(\bm{x}),\tilde{y})-\hat{\mu}|>\epsilon\right)\leqslant \mathrm{e}^{-\mathrm{C_1}(\frac{1}{2}-\mathrm{C_2}\frac{\hat{\sigma}^2}{\epsilon^2})^2}\right.$, where $\mathrm{C_1}$ and $\mathrm{C_2}$ are positive constants.}
 \end{prop}

 In noisy label learning, demonstrating that a loss function is robust involves showing that the estimated loss is equivalent to the loss calculated using clean labels, \emph{i.e.}, for $(\bm{x},\tilde{y})$, $\ell_{\mathrm{RML}}(f(\bm{x}),\tilde{y})$ is close to its true loss $\ell(f(\bm{x}),y)$. However, due to the uncertainty about the correctness of the observed label, it is difficult to know $\ell(f(\bm{x}),y)$. In the proposed RML, we first save samples with the label $\tilde{y}$ in $\tilde{\mathcal{D}}$ as $\tilde{\mathcal{D}}_{\tilde{y}}$, then select a subset $\tilde{\mathcal{D}}^{s}_{\tilde{y}}$ of $\tilde{\mathcal{D}}_{\tilde{y}}$ per Eq. \eqref{prob}. After these two steps, the distribution of $\tilde{\mathcal{D}}^{s}_{\tilde{y}}$ can be approximated as the sample distribution when the true label is $\tilde{y}$, \emph{i.e.}, $\left.P_{\bm{X^s}|\tilde{Y}=\tilde{y}}\rightarrow P_{\bm{X}|Y=\tilde{y}}\right.$. By combining this with $\ell(f(\bm{x}),\tilde{y})$, we obtain $\tilde{\mathcal{D}}_{\tilde{y}}^{*}$ through our analysis. Following this line, the distance between $\hat{\mu}$ and $\ell_{\mathrm{RML}}(f(\bm{x}),\tilde{y})$ can be used to evaluate the robustness of RML. Further explanation of the proposition is attached to Appendix \ref{mexplain}.

\begin{cor}\label{cor1}
    After the loss processing operation in Eq. \eqref{processed}, the upper bound in Prop. \ref{prop2} is reduced.
\end{cor}

Cor. \ref{cor1} shows that our loss processing operation can make the loss estimation more robust. A simple proof can be found in Rmk. \ref{remark} of Appendix \ref{proof}.

\subsection{Combating Label Noise Based on RML}
After the introduction of RML, we then present the common and semi-supervised methods for combating label noise. For the RML-based methods with and without the semi-supervised strategy, we set a traditional training model $f_\theta$ and a momentum model $g_{\theta^{\prime}}$ with parameter as $\theta^{\prime} \in \mathbb{R}^d$ \cite{Tarvainen2017MeanTA} updated following
\begin{equation}
\label{em}
    \theta_{t+1}^{\prime}=(1-\lambda)\times \theta_{t+1}+\lambda \times \theta_{t}^{\prime},
\end{equation}
where $t$ and $\lambda$ are the training step and a weighting parameter. In our experiments, $\lambda$ is set to $0.999$ according to \cite{Tarvainen2017MeanTA}. The pseudo-code of the RML-based method is described in Alg. \ref{semi} of Appendix \ref{algo}.

\paragraph{Common Method Based on RML.} 
As shown in Fig. \ref{fig2}, the traditionally trained model divides training samples into two groups based on their loss values. The left column of the figures shows the prediction results of a model trained using the traditional CE loss, while the right column shows the prediction results of a model trained using RML. Compared to the figures in the left column, those in the right column clearly show that clean and noisy samples are separated into two distinct groups. This demonstrates that RML can prevent the model from learning from noisy samples while still extracting sufficient information from clean samples. By using the knowledge gained from clean samples, we can correct the labels of noisy samples, further improving the model's performance against label noise. As a result, we propose a semi-supervised method.

\paragraph{Semi-supervised Method Based on RML.}
For the semi-supervised method based on RML, we separate noisy and clean samples into two groups.  
Existing selection methods usually adopt several hyperparameters and need to be adjusted according to the noise rate of different training datasets. To achieve an efficient and simple selection approach, we propose a new strategy, which uses $f_\theta$ and $g_{\theta^{\prime}}$ to select samples. For a training sample $(\bm{x},\tilde{y})\in \tilde{\mathcal{D}}$, if its prediction label for both $f_\theta$ and $g_{\theta^{\prime}}$ is equal to $\tilde{y}$, then it is reserved in $\tilde{\mathcal{D}}_{labeled}$. Otherwise, it is stored in $\tilde{\mathcal{D}}_{unlabeled}$.

The semi-supervised model is trained by two steps: common training and semi-supervised training. In common training, $f_\theta$ is trained with RML while $g_{\theta^{\prime}}$ is updated by Eq. \eqref{em}. In semi-supervised training, we use the MixMatch strategy to fully utilize the information of unlabeled samples and further improve the model performance. The detailed procedure can be found in Alg. \ref{semi1} of Appendix \ref{algo}.

\section{Experiments}

\begin{table*}[htb]
\centering
 \small   
\scalebox{1}{
\begin{tabular}{c|c|c|c|c|c|c}
\hline
\multicolumn{1}{c|}{Dataset}  & \multicolumn{3}{c|}{CIFAR-10 } & \multicolumn{3}{c}{CIFAR-100 } \\
\cline{1-7} Method / Symmetric &0.2&0.5&0.8&0.2&0.5&0.8\\ \hline
CE&86.5 $ \pm $ 0.6&80.6 $\pm$ 0.2        &  63.7 $\pm$ 0.8         &    57.9 $\pm$ 0.4          &      47.3 $\pm$ 0.2       &    22.3 $\pm$ 1.2                                        \\
MixUp& 93.2 $\pm$ 0.3  &   88.2 $\pm$ 0.3        &  73.3 $\pm$ 0.3         &    69.5 $\pm$  0.2         &      57.1 $\pm$ 0.6       &    34.1 $\pm$ 0.6                                        \\
DivideMix& 95.6 $\pm$ 0.1  &   94.6 $\pm$ 0.1        &  92.9 $\pm$ 0.3         &    75.3 $\pm$ 0.1          &      72.7 $\pm$ 0.6       &    56.4 $\pm$ 0.3                                        \\
ELR+& 94.9 $\pm$ 0.2  &   93.6 $\pm$ 0.1        &  90.4 $\pm$ 0.2         &    75.5 $\pm$ 0.2          &      71.0 $\pm$ 0.2       &    50.4 $\pm$ 0.8                                        \\
PES& 95.9 $\pm$ 0.1  &   95.1 $\pm$ 0.2        &  93.1 $\pm$ 0.2         &    77.4 $\pm$ 0.3          &      74.3 $\pm$ 0.6       &    61.6 $\pm$ 0.6                                        \\ 
RML-Semi (Ours)& \textbf{ 96.5 $\pm$ 0.2 } &   \textbf{95.7 $\pm$ 0.5}        & \textbf{93.9 $\pm$ 0.2}         &   \textbf{ 78.9 $\pm$ 0.3 }          &     \textbf{ 77.8 $\pm$ 0.7 }      &  \textbf{64.1 $\pm$ 0.3}                                       \\
\hline                   
\end{tabular}
}

\caption{Average test accuracy ($\%$) comparison with state-of-the-art methods with semi-supervised strategy on CIFAR-10
and CIFAR-100. The mean and standard deviation over $5$ runs are presented. Baselines are from \cite{bai2021understanding}. }
\label{semiif}
\end{table*}

\begin{table*}[htb]
\small
\centering
\scalebox{1.0}{
\begin{tabular}{c|c|c|c|c|c|c}
\hline
\multicolumn{1}{c|}{Dataset}  & \multicolumn{3}{c|}{CIFAR-10 } & \multicolumn{3}{c}{CIFAR-100} \\
\cline{1-7} Method / Noise &Instance-0.2&Instance-0.4&Pairflip-0.45&Instance-0.2&Instance-0.4&Pairflip-0.45\\ \hline
CE      & 87.5 $\pm$ 0.5&   78.9 $\pm$ 0.7        &  74.9 $\pm$ 0.7         &   56.8 $\pm$ 0.4          &      48.2 $\pm$ 0.5       &   38.5 $\pm$ 0.6                                        \\
MixUp      &93.3 $\pm$ 0.2  &  87.6 $\pm$ 0.5        &  82.4 $\pm$ 1.0         &   67.1 $\pm$ 0.1          &     55.0 $\pm$ 0.1       &   44.2 $\pm$ 0.5                                       \\
DivideMix      &95.5 $\pm$ 0.1  &  94.5 $\pm$ 0.2        & 85.6 $\pm$ 1.7         &   75.2 $\pm$ 0.2          &     70.9 $\pm$ 0.1       &   48.2 $\pm$ 1.0                                        \\
ELR+      &94.9 $\pm$ 0.1  &  94.3 $\pm$ 0.2        & 86.1 $\pm$ 1.2        &   75.8 $\pm$ 0.1          &     74.3 $\pm$ 0.3       &   65.3 $\pm$ 1.3                                       \\
PES      &95.9 $\pm$ 0.1  &  95.3 $\pm$ 0.1        & 94.5 $\pm$ 0.3         &   77.6 $\pm$ 0.3          &     76.1 $\pm$ 0.4       &   73.6 $\pm$ 1.7                                        \\ 
RML-Semi (Ours)      & \textbf{ 96.3 $\pm$ 0.2 } &   \textbf{95.5 $\pm$ 0.3}        & \textbf{95.2 $\pm$ 0.3}         &   \textbf{ 78.4 $\pm$ 0.2 }          &     \textbf{ 76.8 $\pm$ 0.5 }      &  \textbf{75.9 $\pm$ 0.3}                                       \\
\hline                     
\end{tabular}
}

\caption{Average test accuracy ($\%$) comparison  with state-of-the-art methods with semi-supervised strategy on CIFAR-10
and CIFAR-100. The mean and standard deviation over $5$ runs are presented. Baselines are from \cite{bai2021understanding}.}
\label{semiInsP}
\end{table*}

This section presents the training settings in the first part and then shows the experimental comparison between the proposed method. The last part discusses the impact of some operations in RML on the model performance and analyzes the experimental parameter settings.

\subsection{Experimental Setup}
This section briefly describes some of the experimental settings, including the datasets, network structure, and baselines. Details can be found in Appendix \ref{settings}.

\paragraph{Datasets.} 
We perform experiments on synthetic datasets and real-world datasets. For experiments on synthetic datasets, we choose two commonly used datasets CIFAR-10 and CIFAR-100 with different rates of symmetric label noise, pairflip label noise, and instance-dependent label noise \cite{xia2020part}. The generation of label noise in our experiments follows \cite{bai2021understanding}. For the real-world datasets, we choose Clothing1M and WebVision.

\paragraph{Network Structure.}
All our experiments are performed on Ubuntu 20.04.3 LTS workstations with Intel Xeon 5120 and 5$\times$3090 by PyTorch. To compare with baselines in common experimental results of the model without the semi-supervised strategy, we select ResNet-18 for the CIFAR-10 dataset and ResNet-34 for the CIFAR-100 dataset based on \cite{bai2021understanding}. For semi-supervised experiments, ResNet-18 is used as the backbone for both two datasets. For the real-world datasets, experiments on Clothing1M adopt a pre-trained ResNet-50, while the experiments on WebVision use ResNet-50 trained from scratch.

\paragraph{Baselines.}
We perform two groups of experiments on synthetic and real-world datasets. One trains the model directly, while the other adopts a semi-supervised strategy. For a fair comparison, we compare with different baselines with the same experimental settings. For approaches without semi-supervised strategy, we choose Forward \cite{DBLP:conf/cvpr/PatriniRMNQ17}, Co-teaching \cite{DBLP:conf/nips/HanYYNXHTS18},  Joint-Optim \cite{DBLP:conf/cvpr/TanakaIYA18}, MLNT \cite{DBLP:conf/cvpr/LiWZK19}, T-revision \cite{xia2019anchor}, PCIL \cite{DBLP:conf/cvpr/YiW19}, DMI \cite{xu2019dmi}, and PES \cite{bai2021understanding}. For semi-supervised group, we choose M-correction \cite{DBLP:conf/icml/ArazoOAOM19}, DivideMix \cite{li2020dividemix}, ELR+ \cite{DBLP:conf/nips/LiuNRF20}, Sel-CL \cite{DBLP:conf/cvpr/LiXGL22}, NCR \cite{DBLP:conf/cvpr/IscenVAS22}, and UNICON \cite{karim2022unicon}.

\subsection{Results Comparison on Synthetic Datasets}
We perform two groups of experiments on synthetic datasets. Tab. \ref{unsemisym} shows the experimental comparisons on CIFAR-10 and CIFAR-100 without the semi-supervised strategy. For the experiments on CIFAR-10, we set $k$ to $60$ for a symmetric label noise ratio of $0.8$. For instance-dependent and symmetric label noise with $0.2$ ratio, the $k$ is $600$. The remaining experiments on CIFAR-10 adopt $k=200$. The experiments on CIFAR-100 use $k=50$ when the noise rate is $0.2$. For the experiments on CIFAR-100 with a noise rate of $0.8$, $k$ is $6$. For the rest of the experiments, $k$ is set to $20$. Compared to existing methods, RML helps the model achieve better performance on the two datasets with different noise rates and noise types. RML increases the average test accuracy by about $1\%$ on CIFAR-10 and by about $6\%$ on CIFAR-100. The improvement of the model trained by RML is remarkable when datasets have a higher noise rate, especially on CIFAR-100. Compared with the traditional training model, as shown in Tab. \ref{semiif} and Tab. \ref{semiInsP}, the semi-supervised strategy further improves the model performance against label noise and achieves about $1\%$ average test accuracy improvement on the two datasets, verifying the effectiveness of our proposed RML on synthetic datasets.

\subsection{Results Comparison on Real-world Datasets}

\begin{table}
\setlength\tabcolsep{3pt}
\small
\centering
\scalebox{1.}{
\begin{tabular}{c|c||c|c}
\hline
Method w/o Semi &Accuracy&Method w/ Semi &Accuracy \\ \hline
CE & 69.21 & DivideMix & 74.76 \\  
Joint-Optim &72.16& ELR+ & 74.81\\
DMI & 72.46&PES & 74.99\\ 
MLNT & 73.47&UNICON & 74.98 \\ 
PCIL & 73.49&NCR & 74.6 \\
RML (Ours) &\textbf{73.54}&RML-Semi (Ours) & \textbf{75.14} \\ \hline
\end{tabular}
}

\normalsize
\caption{Average test accuracy (\%) comparison on Clothing1M. Baselines are from \cite{karim2022unicon}.}
\label{clo}

\end{table}

\begin{table*}

\centering
\small
\scalebox{1.0}{
\begin{tabular}{c|c|c||c|c|c}
\hline
Method w/o Semi & WebVision & ILSVRC12& Method w/ Semi& WebVision & ILSVRC12 \\ \hline
Forward & 61.12& 57.36 &DivideMix &77.32&75.20 \\ 
Meta-Net &63.00&57.80&ELR+ & 77.78&70.29\\
Iterative-CV & 65.24&61.60&UNICON & 77.60&75.29  \\
Co-training &63.58&61.48&Sel-CL &79.96 &76.84 \\ 
RML (Ours) &\textbf{73.32} & \textbf{75.04}&RML-Semi(Ours) & \textbf{81.34} & \textbf{77.38} \\ \hline
\end{tabular}
}

\normalsize
\caption{Average test accuracy ($\%$) comparison on WebVision. Baselines are from \cite{bai2021understanding,karim2022unicon}.}
\label{web}
\end{table*}

\begin{table*}
\centering
\small
\scalebox{1.0}{
\begin{tabular}{c|c|c|c|c|c|c}
\hline

\multicolumn{1}{c|}{Dataset}  & \multicolumn{3}{c|}{CIFAR-10 } & \multicolumn{3}{c}{CIFAR-100} \\
\cline{1-7} Method / Noise &Instance-0.4&Symmetric-0.5&Pairflip-0.45&Instance-0.4&Symmetric-0.5&Pairflip-0.45\\ \hline
w/o Loss Processing      & 88.47 $\pm$ 0.37&   86.73 $\pm$ 0.33        &  87.75 $\pm$ 0.71         &   65.43 $\pm$ 0.45          &      62.31 $\pm$ 0.87       &   56.36 $\pm$ 1.23                                        \\
w/o Median Operation   & 85.32 $\pm$ 0.54&   83.61 $\pm$ 0.43        &  85.77 $\pm$ 0.25         &   64.37 $\pm$ 0.56          &      61.81 $\pm$ 0.59       &   55.32 $\pm$ 1.17                                        \\
RML (Ours)      & \textbf{ 90.73 $\pm$ 0.44 } &   \textbf{ 89.13 $\pm$ 0.21}        & \textbf{ 90.14 $\pm$ 0.34}         &   \textbf{ 71.12 $\pm$ 1.12 }          &     \textbf{ 67.05 $\pm$ 0.31 }      &  \textbf{ 64.67 $\pm$ 0.54}                                       \\

\hline 
                     
\end{tabular}
}

\caption{ Ablation study with different training settings. The mean and standard deviation over $5$ runs are presented. }
\label{abl}
\end{table*}

For real-world datasets, it is difficult to obtain noise rates and types. Therefore, methods should be able to handle various types of label noises and uncertain noise rates. To fully evaluate the effect of RML, we perform experiments on two real-world datasets and compare results with baselines.

\paragraph{Performance on Clothing1M.} 
Tab. \ref{clo} compares the model performance with baselines on Clothing1M.  We set $k$ to $200$ for each $1000$ selected training batches. Compared to existing traditionally trained methods, the model trained with RML ahcieves better performance. For the semi-supervised experiments on Clothing1M, the semi-supervised model with RML further strengthens the model performance. Tab. \ref{clo} verifies that the semi-supervised model with RML is able to deal with label noise in such a complex real-world Clothing1M dataset. Compared to the current semi-supervised methods, the semi-supervised model with RML increases the test accuracy by $0.15\%$.

\paragraph{Performance on WebVision.} The WebVision dataset is more challenging because we need to train models from scratch. Due to the unbalanced samples of each class, we select $75\%$ losses of training samples and divide them into $6$ groups to train the model. For the traditional training methods in Tab. \ref{web}, the improvement of the model with RML is remarkable compared to existing methods. Our method achieves an improvement in accuracy of $8\%$ and $14\%$ on WebVision and ILSVRC12, respectively. The comparison between Co-training and RML shows that RML addresses the issues of the existing small-loss criterion and is better at combating label noise in real-world datasets. Tab. \ref{web} demonstrates that the semi-supervised method with RML can protect the model from label noise in real-world datasets and improve the model generalization. The semi-supervised model achieves about $1.5\%$ improvement in test accuracy on WebVision and $0.54\%$ improvement on ILSVRC12, obviously better than the current state-of-the-art methods. 

\subsection{Ablation Studies}

In this part, we discuss the impact of loss processing, median operation and settings of $n$ on experimental results.

\paragraph{Loss Processing.} 
As shown in Tab. \ref{abl}, we compare the experimental results of the model for RML with and without loss processing on CIFAR-10 and CIFAR-100 with various types of label noises, indicating that the proposed loss processing apparently improves the model's performance.

\paragraph{Median Operation.}
The median operation plays an essential role in RML to combat label noise. To verify the function of the median operation, we perform experiments on CIFAR-10 and CIFAR-100. For groups of experiments without the median operation, we simply compute the mean of the selected sample losses. Tab. \ref{abl} shows the experimental results of RML in two cases. Compared with the mean loss, RML with the regroup median operation helps the model achieve better performance, which indicates that the median operation can mitigate the negative impact of selected noisy samples and provide a more appropriate loss estimation for noisy samples than the mean loss.  

\paragraph{Parameter Settings of $n$.} In RML, $n$ is the window size of the median operation and should be an even number. To explore the appropriate setting of $n$, we perform experiments on CIFAR-10 with various types of label noises to evaluate the impact of different values of $n$ on the model convergence speed and performance. As shown in Fig. \ref{fig3} and Tab. \ref{nd} in Appendix, when the value of $n$ is higher, the convergence speed is slower because the higher $n$ is, the smaller the estimated loss is. To take into account the model performance and the convergence speed, we choose $n=6$ on all datasets.


\section{Conclusions}

In this work, we propose RML, a novel method to combat label noise. We analyze the shortcomings of existing methods and introduce the motivation of our proposed method. Then, we present the details of RML and explain some of its main operations. Based on RML, we propose a semi-supervised method based on RML to further improve the model performance against label noise. To verify the effectiveness of RML, we perform a large number of experiments on synthetic datasets with various types of label noises and real-world datasets with unknown noise types and noise rates. These experiments show that RML corrects the distorted losses of noisy samples and provides an appropriate estimation for each training sample. Compared to state-of-the-art methods, these results show that RML further improves the model performance against label noise.

\section*{Acknowledgments}
This work was supported in part by Macau Science and Technology Development Fund under SKLIOTSC-2021-2023, 0072/2020/AMJ and 0022/2022/A1; in part by Research Committee at University of Macau under MYRG2022-00152-FST and  MYRG-GRG2023-00058-FST-UMDF; in part by Natural Science Foundation of China under 61971476; and in part by Alibaba Group through Alibaba Innovative Research Program.

\bibliography{aaai24}

\newpage
\appendix

\section{Algorithms for the RML-based Defense}
\label{algo}

\subsection{Pseudo-code of RML}
In this section, we show the pseudo-codes of the common method and semi-supervised method based on RML in Alg. \ref{semi} and Alg. \ref{semi1}.

\begin{algorithm}[!htbp]
\small
\caption{Regroup Median Loss (RML)}
\label{semi}
\begin{algorithmic}[1]
 \Require Dataset: $\tilde{\mathcal{D}}$; Epoch: $T$; Initial Model: $f_{\theta_{0}}$, $g_{\theta^{\prime}_0}$
	\For{$t=1,\dots,T$}
         \For{$(\bm{x},\tilde{y})$ in $\tilde{\mathcal{D}}$}
         \State $\tilde{\mathcal{D}}_{\tilde{y}}=\big\{(\bm{x}_{i},\tilde{y}_{i})\in\tilde{\mathcal{D}}\mid \tilde{y}_{i}=\tilde{y}\big\}^{m}_{i=1}$
         \State Select $\tilde{\mathcal{D}}_{\tilde{y}}^s$ from $\tilde{\mathcal{D}}_{\tilde{y}}$ per Eq. \eqref{prob}
         \State Calculate $\ell_{\mathrm{RML}}(f_{\theta_{t}}(\bm{x}),\tilde{y})$ per Eq. \eqref{esti} 
        \EndFor
\State Update $\theta_{t}$ by $\ell_{\mathrm{RML}}$
\State Update $\theta_{t}^m$ by Eq. \eqref{em} 
\EndFor
\Ensure Final Model: $f_{\theta_{T}}$, $g_{\theta^{\prime}_T}$
\end{algorithmic}
\end{algorithm}

\begin{algorithm}[!htb]
\small
\caption{The semi-supervised method based on RML}
\label{semi1}

 \begin{algorithmic}[1]
 \Require Dataset: $\left.\tilde{\mathcal{D}}=\{(\bm{x}_i,\tilde{y}_i)\}_{i=1}^N\right.$; Total Training Epoch: $T_1$; Common Training Epoch: $T_2$; Initial Model: $f_{\theta_0}, g_{\theta_0^{\prime}}$
	\For{$t =1,\dots, T_1$}
        \Statex \texttt{\qquad// Common Training} 
        \If{$t \leqslant T_2$}
         \For{$(\bm{x},\tilde{y})$ in $\tilde{\mathcal{D}}$}
         \State Record $\ell^{t}(f_{\theta_{t}}(\bm{x}),\tilde{y})$ as $\mathcal{L}^t(\tilde{\mathcal{D}})$
         \For{$(\bm{x},\tilde{y})$ in $\tilde{\mathcal{D}}_{\tilde{y}}$}
          \State Save $(\bm{x},\tilde{y})$ as $\tilde{\mathcal{D}}_{\tilde{y}}$ and losses as ${\mathcal{L}}^t(\tilde{\mathcal{D}}_{\tilde{y}})$ 
         \State Obtain $\mathcal{W}^t$ and $\ell_{\mathrm{RML}}^t$ for $(\bm{x},\tilde{y})$
        \EndFor
        \EndFor
        \Statex \texttt{\qquad// Semi-supervised Training} 
        \Else
        \For{$(\bm{x},\tilde{y})$ in $\tilde{\mathcal{D}}$}        \If{$f_{\theta_t}(\bm{x})$==$g_{\theta_t^{\prime}}(\bm{x})$==$\tilde{y}$}
        \State Reserve $(\bm{x},\tilde{y})$ in $\tilde{\mathcal{D}}_{labeled}$
        \Else
    \State Reserve $(\bm{x},\tilde{y})$ in $\tilde{\mathcal{D}}_{unlabeled}$
        \EndIf
    \For{$(\bm{x},\tilde{y})$ in $\tilde{\mathcal{D}}_{labeled}$ and $\bm{x}_u$ in $\tilde{\mathcal{D}}_{unlabeled}$}
\State Select a $\gamma \in [0,1]$, $\gamma=\mathrm{max}(\gamma,1-\gamma)$
\State $\ell_{\mathrm{RML}}^t=\ell(f_{\theta_t}(\gamma \bm{x}+(1-\gamma)\bm{x}_u),\tilde{y})$
\EndFor
\EndFor
\EndIf
\State Update $\theta_t$ by $\ell_{\mathrm{RML}}^t$ 
\State Update $\theta_t^{\prime}$ by Eq. \eqref{em}
\EndFor
\Ensure Models: $f_{\theta_{T_1}}, g_{\theta_{T_1}^{\prime}}$
\end{algorithmic}
\end{algorithm}

\subsection{Details for the Loss Update}
\label{details}
Although the procedure obtains an appropriate training sample loss estimate, $\mathcal{L}(\tilde{\mathcal{D}}_{\tilde{y}})$ needs to be changed when $\theta$ is updated in each mini-batch training step, which is extremely time-consuming and limits its application to large-scale datasets. To reduce the time consumption in the model training procedure, we further improve the loss estimation motivated by \cite{DBLP:conf/eccv/JenniF18}. In the improved loss estimation procedure, $\mathcal{L}(\tilde{\mathcal{D}}_{\tilde{y}})$ is updated in each training epoch instead of every backward step. After the $t$-th training epoch, we record all losses of $\tilde{\mathcal{D}}$ as $\mathcal{L}^{t}(\tilde{\mathcal{D}})$ and losses of $\tilde{\mathcal{D}}_{\tilde{y}}$ as $\mathcal{L}^{t}(\tilde{\mathcal{D}}_{\tilde{y}})$. In the $(t+1)$-th training epoch, we select the subset $\tilde{\mathcal{D}}_{\tilde{y}}^s$ from $\tilde{\mathcal{D}}_{\tilde{y}}$ based on sample losses in $\mathcal{L}^{t}(\tilde{\mathcal{D}}_{\tilde{y}})$ according to Eq. \eqref{prob} with loss set as $\mathcal{L}^{t}(\tilde{\mathcal{D}}_{\tilde{y}}^s)$ and compute the mean losses of $n$ subsets of $\tilde{\mathcal{D}}_{\tilde{y}}^s$ as $\mathcal{W}^t$. Then the estimated loss $\ell_{\mathrm{RML}}^t(f(\bm{x}),\tilde{y})$ in the $t$-th epoch can be obtained based on $\mathcal{W}^t$ and $\ell^t(f(\bm{x}),\tilde{y})$ according to Eq. \eqref{esti}. The estimated loss $\ell_{\mathrm{RML}}^{t+1}(f(\bm{x}),\tilde{y})$ based on $\ell_{\mathrm{RML}}^t(f(\bm{x}),\tilde{y})$ and the original loss $\ell^{t+1}(f(\bm{x}),\tilde{y})$ in the $(t+1)$-th epoch is 
\[
    \ell_{\mathrm{RML}}^{t+1}(f(\bm{x}),\tilde{y})=\ell^{t+1}(f(\bm{x}),\tilde{y})\times \frac{\ell_{\mathrm{RML}}^{t}(f(\bm{x}),\tilde{y})}{\ell^t(f(\bm{x}),\tilde{y})}.
\]

In addition, the proposed method may provide a larger estimated loss $\ell_{\mathrm{RML}}(f(\bm{x}),\tilde{y})$ than the original CE loss $\ell(f(\bm{x}),\tilde{y})$ for a training sample, which has the same negative impact as noisy samples. To revise these magnified loss estimates, we set a loss correction condition as 
\[
    \ell_{\mathrm{RML}}(f(\bm{x}),\tilde{y})=
    \begin{dcases}
    \ell_{\mathrm{RML}}(f(\bm{x}),\tilde{y}), & \ell_{\mathrm{RML}}\leqslant \ell, \\ \ell(f(\bm{x}),\tilde{y}), & \ell_{\mathrm{RML}} >\ell.
    \end{dcases}
\]

For a mini-batch $\mathcal{B}$, the mean loss is
\[
    \bar{\ell}_{\mathrm{RML}}=\frac{1}{|\mathcal{B}|}\sum_{(\bm{x},\tilde{y})\in \mathcal{B}}\ell_{\mathrm{RML}}(f(\bm{x}),\tilde{y}).
\]

\begin{figure*}
    \centering
    \includegraphics[scale=0.38]{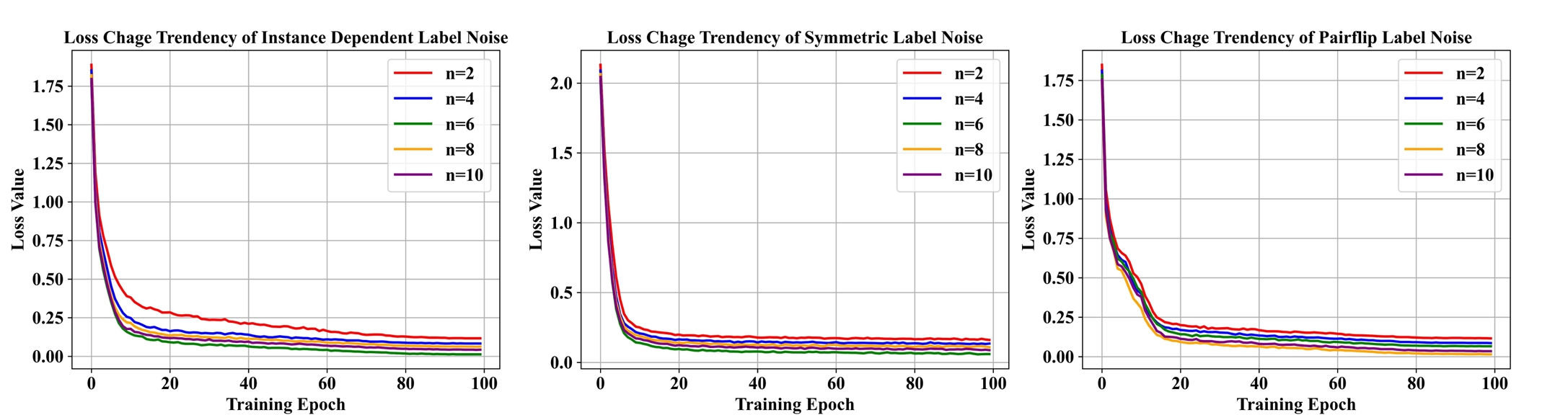}
    \caption{The impact of different $n$ on the  convergence speed of the model training procedure on CIFAR-10 with $0.4$, $0.5$ and $0.45$ ratios of three types of label noises. For experiments on CIFAR-10 with instance-dependent and symmetric label noise, the model converges faster when $n=6$, while the model prefers $n=8$ on CIFAR-10 with pairflip label noise.}
    \label{fig3}
\end{figure*}

\section{Experimental Details}
\label{settings}
\subsection{Datasets} 

In this paper, we perform experiments on synthetic datasets and real-world datasets. For experiments on synthetic datasets, we choose two commonly used datasets, CIFAR-10 and CIFAR-100, with different rates of symmetric label noise, pairflip label noise, and instance-dependent label noise. Both CIFAR-10 and CIFAR-100 contain $50$k training images and $10$k test images of size $32 \times 32$. According to some existing works \cite{xia2020part,cui2022improving}, symmetric label noise is generated by uniformly flipping labels for a percentage of the training dataset to all possible labels. Pairflip label noise is generated by transforming samples to their adjacent classes. 

For instance-dependent label noise, the samples are flipped to classes with the same features for them. To compare with baselines, we perform experiments on the two datasets with the different types of label noises and various noisy rates following previous work \cite{bai2021understanding}. For the real-world datasets, we choose two commonly used datasets: Clothing1M \cite{xiao2015learning} and WebVision \cite{li2017WebVision}.  Clothing1M is a large-scale dataset with real-world noisy labels, whose images are clawed from online shopping websites and whose labels are generated based on surrounding texts. It contains $1$ million training images, $15$k validation images, and $10$k test images with clean labels. WebVision contains $2.4$ million images (obtained from Flickr and Google) categorized into the same $1,000$ classes as in ImageNet ILSVRC12. Following the previous studies \cite{karim2022unicon}, we use the first $50$ classes of the Google image subset as training data.

\subsection{Parameter Settings}

All of our experiments are performed on Ubuntu 20.04.3 LTS workstation with Intel Xeon 5120 and $5 \times 3090$ by PyTorch. To compare with baselines in common experimental results, we select ResNet-18 for the CIFAR-10 dataset and ResNet-34 for the CIFAR-100 dataset based on \cite{bai2021understanding}. The selected models for these two datasets are trained for $200$ epochs and SGD optimizer with momentum of $0.9$ and weight decay of $5\times 10^{-4}$. The initial learning rate is set to $0.1$ and adjusted by cosine annealing from $0.1$ to $0.0001$. In semi-supervised result comparison, ResNet-18 is used for both two datasets. The initial learning rate is $0.02$ for SGD optimizer with momentum of $0.9$ and weight decay of $5\times 10^{-4}$. For semi-supervised training, the model needs to be trained for $400$ epochs in total. The model is trained directly for $50$ epochs on CIFAR-10 and for $35$ epochs on CIFAR-100 and put on semi-supervised training for the remaining training epochs. In the training procedure, we set the batch size to $128$ with the image size $32\times 32$. In addition, the MixMatch parameter settings of the semi-supervised model follow \cite{li2020dividemix, DBLP:conf/ijcai/GuiWT21}.

For the experiments on real-world datasets, we adopt a pre-trained ResNet-50 as the backbone for Clothing1M and an initial ResNet-50 for WebVision. For these two datasets, the model is optimized by SGD with momentum as $0.9$ and weight decay as $5\times 10^{-4}$. The initial learning rate is $2\times 10^{-3}$ for Clothing1M and $1\times 10^{-2}$ for the WebVision dataset. All these learning rates are adjusted by cosine annealing limited as $2\times 10^{5}$. For the semi-supervised training, experiments are performed in $200$ and $300$ epochs for the Clothing1M and WebVision, respectively. For Clothing1M, we set the batch size to $32$. The image is augmented by a random crop size $224\times 224$ from $256\times 256$. In the common training procedure, we select $1000$ batches of samples each epoch as \cite{li2020dividemix} for training. During the semi-supervised model training procedure, the model uses $10$ epochs for common training and the rest of the epochs for semi-supervised training. For the training procedure on WebVision, we set the batch size as $32$. The images are augmented by a random crop size $224\times 224$ from the original size $320\times 320$. Experiments on WebVision take $200$ epochs to perform the common training step and the rest of the training epochs for the semi-supervised training. To show the robustness of RML, we set $n$ to $6$ for all datasets. The setting of $k$ needs to be changed with the noise rate.

\subsection{More Experimental Results}
\label{eer}
This section shows the experimental results of memory and time consumption in Tab. \ref{ts} and parameter settings of $n$ on CIFAR-10 in Fig. \ref{fig3} and Tab. \ref{nd}.

\begin{table}
\centering
\small
\scalebox{1.0}{
\begin{tabular}{ccc}
\toprule
Method & Time & Memory \\ 
\midrule
CE  & 1.0 & 2.2 \\
T-revision & 3.9 & 2.5\\
ELR+ & 2.5 & 2.6\\
PES & 3.6 & 2.5 \\
RML (Ours) & 2.4 & 2.7 \\
\midrule
DivideMix & 6.3 & 3.6\\
RML-semi (Ours) & 5.0 & 2.8\\
\bottomrule
\end{tabular}
}
\caption{Comparison of training time (hrs) and memory space (GB) on CIFAR-10 with 50\% symmetric label noise.}
\label{ts}
\end{table}

\begin{table}
\centering
\small
\scalebox{1.0}{
\begin{tabular}{cccc}
\toprule
$n$ &Instance & Symmetric & Pairflip  \\ \midrule
2 & 87.41$\pm$0.86& 85.39$\pm$0.93 &87.65$\pm$0.81\\ 
4 & 90.29$\pm$0.24&88.87$\pm$0.39 &88.94$\pm$0.57\\
6 & \textbf{90.73$\pm$0.44}&\textbf{89.13$\pm$0.21} &89.67$\pm$0.49\\
8 &90.22$\pm$0.73&88.76$\pm$0.31 &\textbf{90.14$\pm$0.34}\\
10 &89.71$\pm$0.23&88.69$\pm$0.15 &89.43$\pm$0.24\\ 
\bottomrule 
\end{tabular}
}

\caption{Ablation study of $n$ on CIFAR-10.}
\label{nd}
\end{table}

\section{Details of Proposition \ref{prop2}}

\subsection{Proof}
\label{proof}

\begin{proof}
For convenience, we denote $\ell_{\mathrm{RML}}(f(\bm{x}),\tilde{y})$ by $\ell_{\mathrm{RML}}$. First of all, observe that the event
 \[
     \left\{|\ell_{\mathrm{RML}}-\hat{\mu}|>\epsilon\right\}
 \]
implies that at least $\frac{n+1}{2}$ of mean loss $\ell^s_i$ in $\mathcal{W}^{*}$ have to be outside $\epsilon$ distance to $\hat{\mu}$. Namely,
\[
    \left\{|\ell_{\mathrm{RML}}-\hat{\mu}|>\epsilon\right\}\subset\left\{\sum^{n+1}_{i=1} \mathds{1}(|\ell^s_i-\hat{\mu}|>\epsilon)\geqslant\frac{n+1}{2}\right\}.
\]

Define $Z_i=\mathds{1}(|\ell^s_i-\hat{\mu}|>\epsilon)$. Let $p_{\epsilon,k}=\mathbb{E}[Z_i]=\mathbb{P}\left(|\ell^s_i-\hat{\mu}|>\epsilon\right)$, $i=1,\dots,n$, and $p_{\epsilon}=\mathbb{E}[Z_{n+1}]=\mathbb{P}\left(|\ell-\hat{\mu}|>\epsilon\right)$, then we can obtain
\[
    \begin{split}
        &\mathbb{P}\left(|\ell_\mathrm{RML}-\hat{\mu}|>\epsilon\right) \leqslant  \mathbb{P}\left(\sum^{n+1}_{i=1} Z_i \geqslant\frac{n+1}{2}\right)\\
        ={}&\mathbb{P}\left(\sum^{n+1}_{i=1} (Z_i-\mathbb{E}[Z_i]) \geqslant\frac{n+1}{2}-n p_{\epsilon,k}-p_{\epsilon}\right)\\
        ={}&\mathbb{P}\left(\frac{1}{n+1}\sum^{n+1}_{i=1} (Z_i-\mathbb{E}[Z_i]) \geqslant\frac{1}{2}-\frac{n}{n+1}p_{\epsilon,k}-\frac{1}{n+1}p_{\epsilon}\right).
    \end{split}
\]

Because $Z_i$ are independent random variable bounded by $1$, then by Hoeffding's inequality (one-sided), with any $\tau>0$, we can obtain
\[
    \mathbb{P}\left(\frac{1}{n+1}\sum^{n+1}_{i=1} (Z_i-\mathbb{E}[Z_i]) \geqslant \tau\right)\leqslant \mathrm{e}^{-2(n+1)\tau^2}.
\]

As a result,
\[
    \begin{split}
        & \mathbb{P}\left(|\ell_{\mathrm{RML}}-\hat{\mu}|>\epsilon\right)\\
        \leqslant{} & \mathbb{P}\left(\frac{1}{n+1}\sum^{n+1}_{i=1} (Z_i-\mathbb{E}[Z_i]) \geqslant \frac{1}{2}-\frac{n}{n+1}p_{\epsilon,k}-\frac{1}{n+1}p_{\epsilon}\right) \\
        \leqslant{} & \mathrm{e}^{-2(n+1)(\frac{1}{2}-\frac{n}{n+1}p_{\epsilon,k}-\frac{1}{n+1}p_{\epsilon})^2}.
    \end{split}
\]

Due to $\mathrm{Var}_{\bm{X^*}|\tilde{Y}=\tilde{y}}[\ell(f(\bm{X^{*}}),\tilde{y})]=\hat{\sigma}^2<\infty$ and Chebyshev’s inequality, we can obtain
\[
\begin{split}
    p_{\epsilon,k}&=\mathbb{P}(|\ell^s_i-\hat{\mu}|>\epsilon)\leqslant \frac{\hat{\sigma}^2}{k\epsilon^2},\\
     p_{\epsilon}&=\mathbb{P}(|\ell-\hat{\mu}|>\epsilon)\leqslant \frac{\hat{\sigma}^2}{\epsilon^2}.
\end{split}
\]

Then the bound is
\[
\begin{split}
     \mathbb{P}\left(|\ell_{\mathrm{RML}}-\hat{\mu}|>\epsilon\right)&\leqslant \mathrm{e}^{-2(n+1)(\frac{1}{2}-\frac{n}{n+1}p_{\epsilon,k}-\frac{1}{n+1}p_{\epsilon})^2}\\ 
     &\leqslant \mathrm{e}^{-2(n+1)(\frac{1}{2}-\frac{n}{n+1}\frac{\hat{\sigma}^2}{k\epsilon^2}-\frac{1}{n+1}\frac{\hat{\sigma}^2}{\epsilon^2})^2}\\
     &=\mathrm{e}^{-2(n+1)(\frac{1}{2}-\frac{n+k}{n+1}\frac{\hat{\sigma}^2}{k\epsilon^2})^2}\\
     &=\mathrm{e}^{-\mathrm{C_1}(\frac{1}{2}-\mathrm{C_2}\frac{\hat{\sigma}^2}{\epsilon^2})^2},
\end{split}
\]
where $\mathrm{C_1}=2(n+1),\mathrm{C_2}=\frac{n+k}{k(n+1)}>0.$
\end{proof}

\begin{remark}
\label{remark}
    Note that after the operation in Eq. \eqref{processed}, intuitively, losses are closer to the true loss and so that number of $\ell_i^s$ outside $\epsilon$ distance to $\hat{\mu}$ decreases. Therefore,  $\sum^{n+1}_{i=1} \mathds{1}\left(|\ell^s_i-\hat{\mu}|>\epsilon\right)$ decreases and so does $\mathbb{P}\left(\sum^{n+1}_{i=1} Z_i \geqslant\frac{n+1}{2}\right)$, i.e., the upper bound in Prop. \ref{prop2} is reduced.
\end{remark}

\subsection{More Explanation}
\label{mexplain}
Prop. \ref{prop2} shows that the probability that the distance between $\hat{\mu}$ and the estimated $\ell_{\mathrm{RML}}(f(\bm{x}),\tilde{y})$ is greater than $\epsilon$ is very small. It proves that the loss estimated by RML is robust and has a stable upper bound if less than half of the $k$-sample subsets are affected by distorted sample losses. The reason is that the mean losses of $k$-sample subsets of $\tilde{\mathcal{D}}_{\tilde{y}}$ without noisy samples are reliable estimates of $\ell(f(\bm{x}),\tilde{y})$ and have no negative impact on the estimated $\ell(f(\bm{x}),\tilde{y})$. In the training procedure, the proposed loss processing operation in Eq. \eqref{processed} reduces the selection probability of noisy samples in $\tilde{\mathcal{D}}_{\tilde{y}}^{*}$. Moreover, $n\times k$ can be adjusted to reduce the number of noisy samples selected in $\tilde{\mathcal{D}}_{\tilde{y}}^{*}$ and make less than half of $k$-sample subsets affected by noisy samples. Improving $n$ can also reduce the negative impact of noisy samples on the estimated loss. Then, the robustness of the estimated loss by RML can be confirmed. Based on the above statement, using the median of $k$-sample subset mean losses can achieve stable and robust loss estimation.

\end{document}